\pdfoutput=1

\documentclass[11pt]{article}

\usepackage{EMNLP2022}

\usepackage{color}

\usepackage{times}
\usepackage{latexsym}
\usepackage{listings}
\usepackage{xcolor}
\usepackage{color}
\usepackage{latexsym}
\usepackage{graphicx}
\usepackage{booktabs}
\usepackage{algorithm}
\usepackage{algorithmic}
\usepackage{caption}
\usepackage{subfigure}
\usepackage{multirow}
\usepackage{microtype}
\usepackage{amssymb}

\usepackage{footmisc}

\usepackage[T1]{fontenc}

\usepackage[utf8]{inputenc}

\usepackage{microtype}

\usepackage{inconsolata}

\usepackage{microtype}
\usepackage{amsmath, amssymb}
\usepackage{booktabs}
\usepackage{graphicx}
\usepackage{multirow}
\usepackage{xpinyin}

\usepackage{bm}
\usepackage{algorithm}
\usepackage{pifont}
\usepackage{subfigure}
\usepackage{xspace}
\usepackage{float}
\usepackage{graphicx}

\title{\modelname: Complex Dialogue Utterance Splitting and Reformulation \\for Multiple Intent Detection}

\newcommand{\modelname}{DialogUSR\xspace}

\definecolor{mygreen}{rgb}{0.00,0.70,0.5}
\makeatletter

\newcommand{\Rmnum}[1]{\expandafter\@slowromancap\romannumeral #1@}
\makeatother

\author{
Haoran Meng$^1$$\footnotemark[1]$ \quad 
Xin Zheng$^2$ $^4$$\footnotemark[1]$ \quad 
Tianyu Liu$^3$$\footnotemark[1]$ $\footnotemark[2]$ \quad Zizhen Wang$^3$ \quad He Feng$^3$ \\ 
\textbf{Binghuai Lin$^3$ \quad Xuemin Zhao$^3$ \quad Yunbo Cao$^3$ \quad Zhifang Sui$^1$$\footnotemark[2]$} \\ 
$^1$ MOE Key Laboratory of Computational Linguistics, Peking University, China \\
$^2$ Institute of Software, Chinese Academy of Sciences, China \\
$^3$ Tencent Cloud Xiaowei 
$^4 $University of Chinese Academy of Sciences, China \\
\texttt{haoran@stu.pku.edu.cn;zhengxin2020@iscas.ac.cn;\{rogertyliu,zizhenwang,}\\
\texttt{mobisysfeng,binghuailin,xueminzhao,yunbocao\}@tencent.com;szf@pku.edu.cn}
}

\begin{document}

\maketitle

\renewcommand{\thefootnote}{\fnsymbol{footnote}}

\begin{abstract}
While interacting with chatbots, users may elicit multiple intents in a single dialogue utterance.
Instead of training a dedicated multi-intent detection model, we propose \modelname, a dialogue utterance splitting and reformulation task 
that first splits multi-intent user query into several single-intent sub-queries and then recovers all the coreferred and omitted information in the sub-queries.
DialogUSR can serve as a plug-in and domain-agnostic module that empowers the multi-intent detection for the deployed chatbots with minimal efforts.
We collect a high-quality naturally occurring dataset that covers 23 domains with a multi-step crowd-souring procedure.
To benchmark the proposed dataset, we propose multiple action-based generative models that involve end-to-end and two-stage training, and conduct in-depth analyses on the pros and cons of the proposed baselines.

\end{abstract}

\footnotetext[1]{Equal contribution.}
\footnotetext[2]{Corresponding authors.}

\renewcommand{\thefootnote}{\arabic{footnote}}

\section{Introduction}
\begin{figure}[t]
    \centering
    \includegraphics[width=0.95\linewidth]{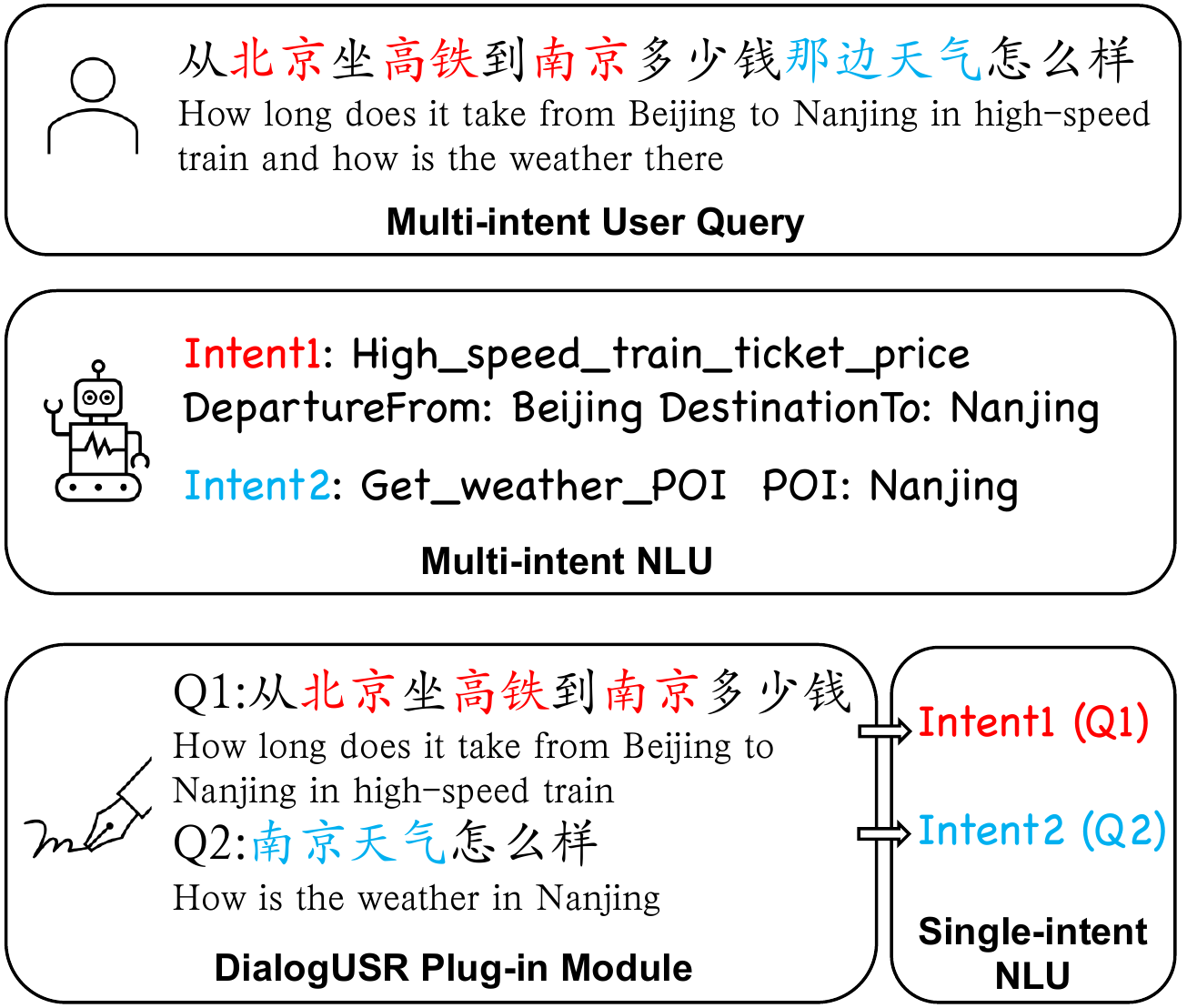}    
    \caption{The task illustration for \modelname. It serves as a plug-in module that empowers multi-intent detection capability for deployed single-intent NLU systems.}
    \label{fig:task_intro}
\end{figure}

Thanks to the technological advances of natural language processing (NLP) in the last decade, modern personal virtual assistants like Apple Siri, Amazon Alexa have managed to interact with end users in a more natural and human-like way. Taking chatbots as human listeners, users may elicit multiple intents within a single query. For example, in Figure \ref{fig:task_intro}, a single user query triggers the inquiries on both high-speed train ticket price and the weather of destination.
To handle multi-intent user queries, a straightforward solution is to train a dedicated natural language understanding (NLU) system for multi-intent detection. 
\citet{Rychalska2018MultiIntentHN} first adopted hierarchical structures to identify multiple user intents. \citet{gangadharaiah-narayanaswamy-2019-joint}  explored the joint multi-intent and slot-filling task with a recurrent neural network. \citet{qin-etal-2020-agif} further proposed an adaptive graph attention network to model the joint intent-slot interaction.
To integrate the multi-intent detection model into a product dialogue system, the developers would make extra efforts in continuous deployment, i.e. technical support for both single-intent and multi-intent detection models, and system modifications, i.e. changes in the APIs and implementations of NLU and other related modules.

To provide an alternative way towards understanding multi-intent user queries, we propose complex dialogue utterance splitting and reformulation (DialogUSR) task with corresponding benchmark dataset that firstly splits the multi-intent query into several single-intent sub-queries and then recover the coreferred and omitted information in the sub-queries, as illustrated in Fig \ref{fig:task_intro}. With the proposed task and dataset, the practitioners can train a multi-intent query rewriting model that serves as a plug-in module for the existing chatbot system with minimal efforts. The trained transformation models are also domain-agnostic in the sense that the learned query splitting and rewriting skills in DialogUSR are generic for multi-intent complex user queries from diverse domains.

We employ a multi-step crowdsourcing procedure to annotate the dataset for DialogUSR which covers 23 domains with 11.6k instances. The naturally occurring coreferences and omissions account for 62.5\% of the total human-written sub-queries, which conforms to the genuine user preferences. Specifically we first collect initial queries from 2 Chinese task-oriented NLU datasets that cover real-world user-agent interactions, then ask the annotators to write the subsequent queries as they were sending multiple intents to the chatbots, finally we aggregate the human written sub-queries and provide completed sub-queries if coreferences and omissions are involved. We also employ multiple screening and post-checking protocols in the entire data creation process, in order to ensure the high quality of the proposed dataset.

For baseline models, we carefully analyze the transformation from the input multi-intent queries to the corresponding single-intent sub-queries and summarize multiple rewriting actions, including \texttt{deletion}, \texttt{splitting}, \texttt{completion} and \texttt{causal completion} which are the local edits in the generation. Based on the summarized actions, we proposed three types of generative baselines: end-to-end, two-stage and causal two-stage models which are empowered by strong pretrained models, and conduct a series of empirical studies including the exploration on the best action combination, the model performance on different training data scale and existing multi-intent NLU datasets.

We summarize our contributions as follows\footnote{Code and data are provided in \url{https://github.com/MrZhengXin/multi_intent_2022}.}:

\textbf{1)} The biggest challenges of multi-intent detection (MID) in the deployment is the heavy code refactoring on a running dialogue system which already does a good job in single-intent detection. It motivates us to design DialogUSR, which serves as a plug-in module and eases the difficulties of incremental development.

\textbf{2)} Prior work on MID has higher cost of data annotation and struggles in the open-domain or domain transfer scenarios. Only NLU experts can adequately annotate the intent/slot info for a MID user query, and the outputs of MID NLU models are naturally limited by the pre-defined intent/slot ontology. In contrast, DialogUSR datasets can be easily annotated by non-experts, and the derived models are domain-agnostic in the sense that the learned query splitting, coreference/omission recovery skills are generic for distinct domains

\textbf{3)} Presumably MID is more difficult than single intent detection (SID) given the same intent/slot ontology. From the perspective of task (re)formulation, DialogUSR is the first to convert a MID task to multiple SID tasks (the philosophy  of 'divide and conquer') with a relatively low error propagation rate, providing an alternative and effective way to handle the MID task.

\begin{figure}[t!] 
    \centering 
    \includegraphics[width=1.0\linewidth]{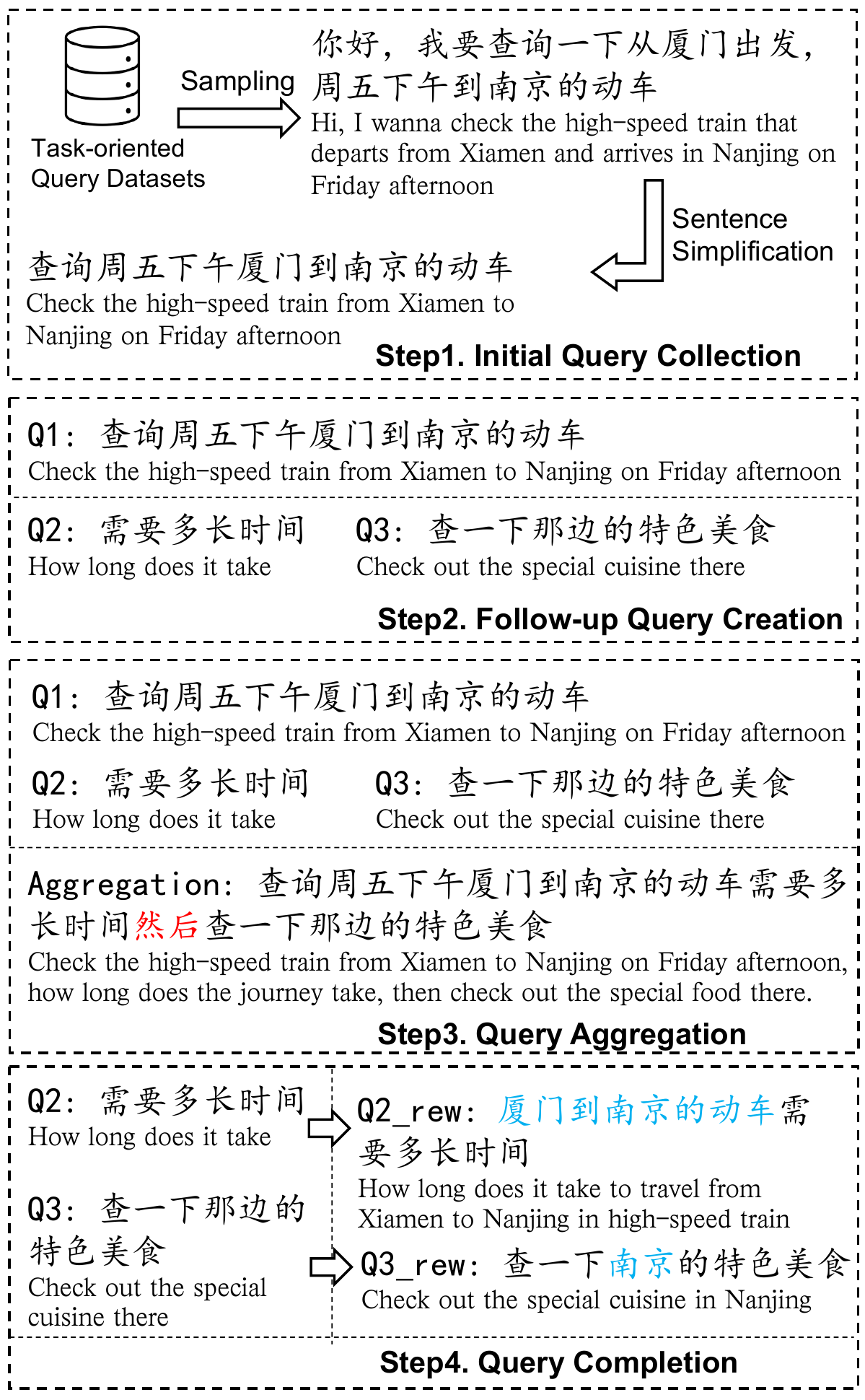}
    \caption{The overview for the data collection procedure of DialogUSR. Firstly we sample initial queries from task-oriented NLU datasets (Sec. \ref{sec:initial_query}), then we hire crowdsource workers to write follow-up queries (Sec. \ref{sec:followup_query}). To aggregate the annotated queries, we propose text filler templates (marked in red, Sec. \ref{sec:query_aggregation}) and post-processing procedure. Finally we ask annotators to recover the missing information in the incomplete utterances (marked in blue, Sec. \ref{sec:query_completion}). } 
    \label{fig:data_overview}
\end{figure}

\section{Dataset Creation}
\label{sec:dataset_creation}
We collect a high quality dataset via a 4-step crowdsourcing procedure as illustrated in Fig \ref{fig:data_overview}.

\subsection{Initial Query Collection}
\label{sec:initial_query}
In order to determine the topic of the multi-intent user query, 
we sample an initial query from two Chinese user query understanding datasets for task-oriented conversational agents, namely SMP-ECDT\footnote{\url{http://ir.hit.edu.cn/SMP2017-ECDT}}\citep{zhang2017first} and RiSAWOZ\footnote{\url{https://github.com/terryqj0107/RiSAWOZ}} ~\citep{quan-etal-2020-risawoz}. 
Then we ask human annotators to simplify the initial queries that have excessive length (longer than 15 characters), or are too verbose or repetitive in terms of semantics\footnote{The sentence simplification phase makes the annotated multi-intent queries sound more natural, as users are not likely to elicit a lengthy query. Given the fact that we would add 2 or 3 following sub-queries to the initial queries, they should be simplified to keep a proper query length (Fig \ref{fig:data_overview}).}. 
RiSAWOZ is a a large-scale multi-domain Chinese Wizard-of-Oz NLU dataset with rich semantic annotations, which covers 12 domains in \emph{tourist attraction}, \emph{railway}, \emph{hotel}, \emph{restaurant}, etc. SMP-ECDT is released as the benchmark for the ``domain and intent identification for user query'' task in the evaluation track of Chinese Social Media Processing conference (SMP) 2017 and 2019. It covers divergent practical user queries from 30 domains which are collected from the production chatbots of iFLYTEK.
We use the two source datasets as our query resources as they comprise a variety of common and naturally occurring user queries in daily life for task-oriented chatbot and cover diverse domains and topics.

\subsection{Follow-up Query Creation}
\label{sec:followup_query}
After specifying an initial query, we ask human annotators to put themselves in the same position of a real end user and imagine they are eliciting multiple intents in a single complex user query while interacting with conversational agents.
The annotators are instructed to write up to 3 subsequent queries on what they need or what they would like to know about according to the designated initial query. 
Although most subsequent queries stick to the topic of the initial query, we allow the human annotators to switch to a different topic which is unrelated to the initial query\footnote{In fact, we neither encourage nor discourage topic switching in the annotation instruction.}. For example in Figure \ref{fig:task_intro}, the second sub-query asks about the weather in Nanjing, where the initial query is an inquiry on the railway information.
We observe that 37.3\% annotated multi-intent queries involve topic switching by manually checking 300 subsampled instances in the training set, which conforms to the user behaviour in the real-world multi-intent queries.

\begin{figure*}[t]
    \centering
    \includegraphics[width=0.95\linewidth]{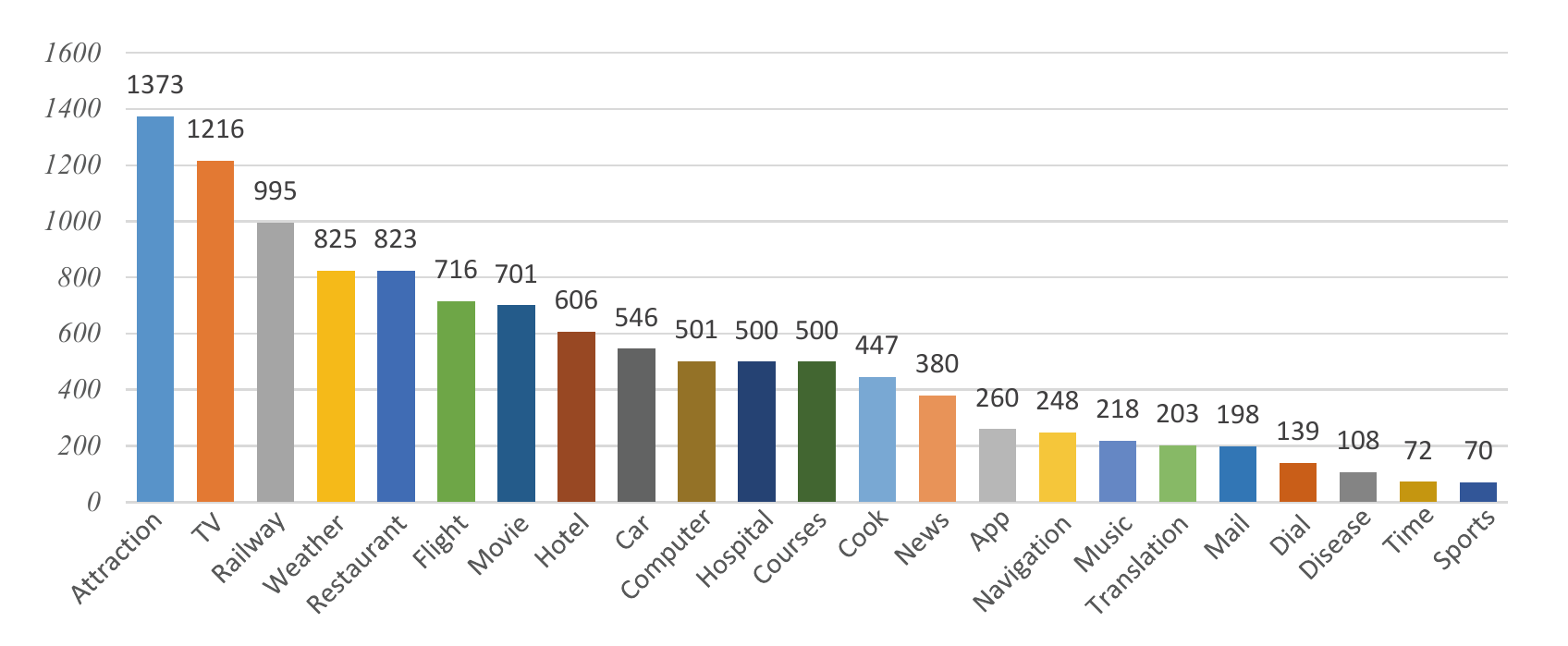}    
    \caption{The domain statistics of DialogUSR, which covers diverse domains in the conversational agents.
    }
    \label{fig:domain_statistics}
\end{figure*}

\subsection{Query Aggregation}
\label{sec:query_aggregation}
In the pilot study, we tried to ask human annotators to manually aggregate the sub-queries but found that the derived queries are somewhat lack of variations in the conjunctions between the sub-queries, as the annotators tend to always pick up the most common Chinese conjunctions like 'and', 'or', 'then'. We even observed sloppy annotators trying to hack the annotation job by not using any conjunctions at all for each query (most queries are fluent even without conjunctions). In a nutshell, we find it challenging to screen the annotators and ensure the diversity and naturalness of the derived query in the human-only annotation.
We then resort to human-in-the-loop annotation, sampling from a rich conjunction set to connect sub-queries and post-checking the sentence fluency of aggregated queries by GPT-2. After each round of annotation (we have 6 rounds of annotations), we randomly pick up 100 samples and check their quality, finding that over 95\% of samples are of high quality. Actually most sentences in the Fig 9 (appendix) are fluent and natural (especially in Chinese) without cherry-picking.

More concretely we propose a set of pre-defined templates that correspond to different text infilling strategies between consecutive queries. 
Specifically, with a 50\% chance we concatenate two consecutive queries without using any text filler. For the other 50\% chance, we sample a piece of text from a set of pre-defined text fillers with different sampling weights, such as 
``\begin{CJK*}{UTF8}{gbsn}首先\end{CJK*}'' (first of all), 
``\begin{CJK*}{UTF8}{gbsn}以及\end{CJK*}'' (and), ``\begin{CJK*}{UTF8}{gbsn}我还想知道\end{CJK*}'' (I also would like to know), 
``\begin{CJK*}{UTF8}{gbsn}接下来\end{CJK*}'' (then),
``\begin{CJK*}{UTF8}{gbsn}最后\end{CJK*}'' (finally), 
and then use the sampled text filler as a conjunction while concatenating consecutive queries.
Although being locally coherent, the derived multi-intent query may still exhibit some global incoherence and syntactic issues, especially for longer text.
We thus post-process the derived query with a ranking procedure as an additional screening step. For each annotated query set, we generate 10  candidate multi-intent queries with different sampled templates and rank them according to language model perplexity using a GPT-2 (117M) model. We only keep the the candidate with lowest perplexity to ensure the fluency and syntactic correctness. 
To avoid trivial hacks in the complex query splitting, we remove all the punctuations in the aggregated query, which conforms to the default settings of most production chatbots, i.e. no punctuations in the spoken language understanding phase after going through the automatic speech recognition module.

\begin{figure*}[t]
    \centering
    \includegraphics[width=0.95\linewidth]{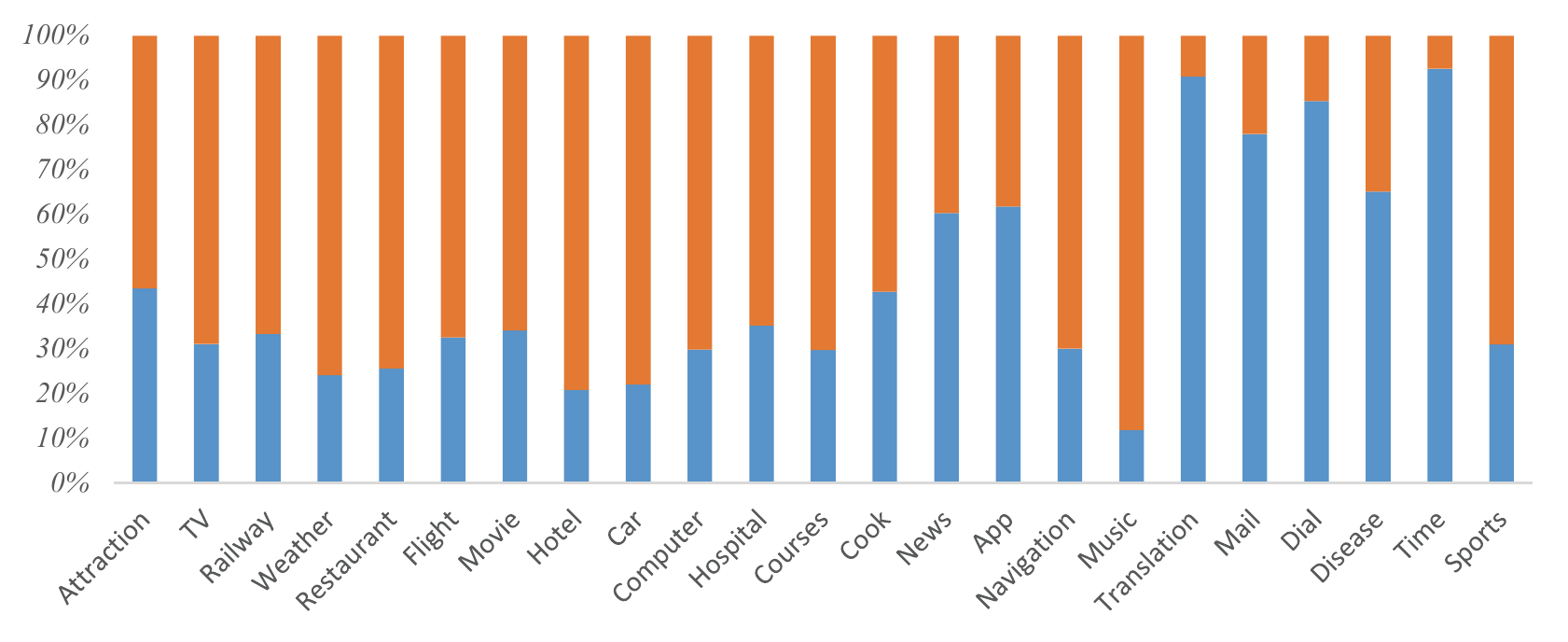}    
    \caption{The ratio of the incomplete utterances in the gold outputs of DialogUSR. The blue bar signifies incomplete utterances which requires rewriting while the orange bar represents complete utterances. 
    }
    \label{fig:type_statistics}
\end{figure*}

\subsection{Query Completion}
\label{sec:query_completion}
After assembling the multi-intent user queries, we observe that incomplete utterances, such as co-references and omissions, are frequently occurring which account for 62.5\% of total human-written subsequent queries. Note that, in the annotation instruction, we do not explicitly ask the crowdsource worker to use coreferences or omissions while writing the subsequent queries in the \emph{follow-up query creation} phase. 
The naturally occurring incomplete utterances reflect genuine user preferences while sending out multiple intents. 
To gather sufficient information while splitting multi-intent queries into independent single-intent queries, we ask another group of annotators\footnote{The \emph{query completion} phase starts when \emph{follow-up query creation} phase has finished. We hire another group of annotators that did not participate in the follow-up query writing task to screen the quality of rewritten queries while doing \emph{query completion}.}
to write the completed utterances by recovering omitted and co-referred information for the incomplete queries.

\subsection{Data Annotation Settings}
To perform human annotation, we hired crowdsource workers from an internal data annotating group. The workers were limited to those who have abundant hand-on experiences in annotating conversational data with good records (recognized as experts in the internal assessment, rejection rate $\leq$ 1\%).
Additionally, all the workers were screened via a 10-case qualification test that covers various annotation tasks in Sec \ref{sec:initial_query} to Sec \ref{sec:query_completion} (correctly annotating 8 out of 10 cases).
They were paid 0.6\$ per datapoint, which is more than prevailing local minimum wage.
We split the entire annotation procedure into multiple rounds and hire another group of human judges to post-check the quality of annotated dataset and filter unqualified instances after each round. In this way, we create a high-quality crowdsourcing dataset.

\section{Dataset Analysis}
\label{sec:data_analysis}
\paragraph{Dataset Statistics}

In total, after accumulating annotations for several rounds, we obtain 11,669 instances. We conduct 6 rounds of annotation, increasing the annotation scale with each round (ranging from $\sim$100 instances/round to $\sim$4000 instances/round). On average, an aggregated multi-intent complex query from the proposed DialogUSR dataset comprises 36.7 Chinese characters by assembling 3.6 single-intent queries (including initial and follow-up queries). After recovering missing information in the query completion phase (Sec \ref{sec:query_completion}), the average lengths of completed initial query, first follow-up query, second follow-up query and 
third follow-up query are 11.9, 12.3, 12.4, 10.8 respectively. 
We split the dataset into train, validation and test sets with sizes of 10,169 , 500, 1,000 respectively.

\paragraph{Domain Statistics}
The domain statistics of DialogUSR is depicted in Fig \ref{fig:domain_statistics}.
Thanks to the diverse domains of our source datasets, DialogUSR covers 23 domains that chatbot users frequently query on in their daily life.
Additionally, as mentioned in Sec \ref{sec:followup_query}, the annotators proactively switch topics or domains in the data creation procedure. We find that, on average, a complex query in DialogUSR involves 1.4 domains, showing the potential usage of recognizing user intents across different domains.
The models training on the DialogUSR dataset can deal with divergent situations in the practical usage while accommodating the utility of personal virtual assistant.

\paragraph{Incomplete Utterance Analysis}
Existing multi-intent detection datasets, such as MixATIS and MixSNIPS \cite{qin-etal-2020-agif}, were created using simple heuristic rules, e.g. adding a particular conjunction ``and'' while concatenating two single-intent queries. The simple heuristic datasets largely undermine the multi-intent detection in the real-world conversational agents, where users naturally interact with chatbots with coreferences and omissions.
As highlighted in Sec \ref{sec:query_completion}, nearly two thirds of human-written subsequent queries are incomplete. We further show the incomplete ratio of follow-up queries for different domains in Fig \ref{fig:type_statistics}.
In the incomplete utterances, according to our statistics, only 2.4\% of them belong to the coreferred phenomenon, showing that users prefer not using pronouns to refer to previously mentioned entities. 

\begin{figure}[t] 
    \centering 
    \includegraphics[width=1.0\linewidth]{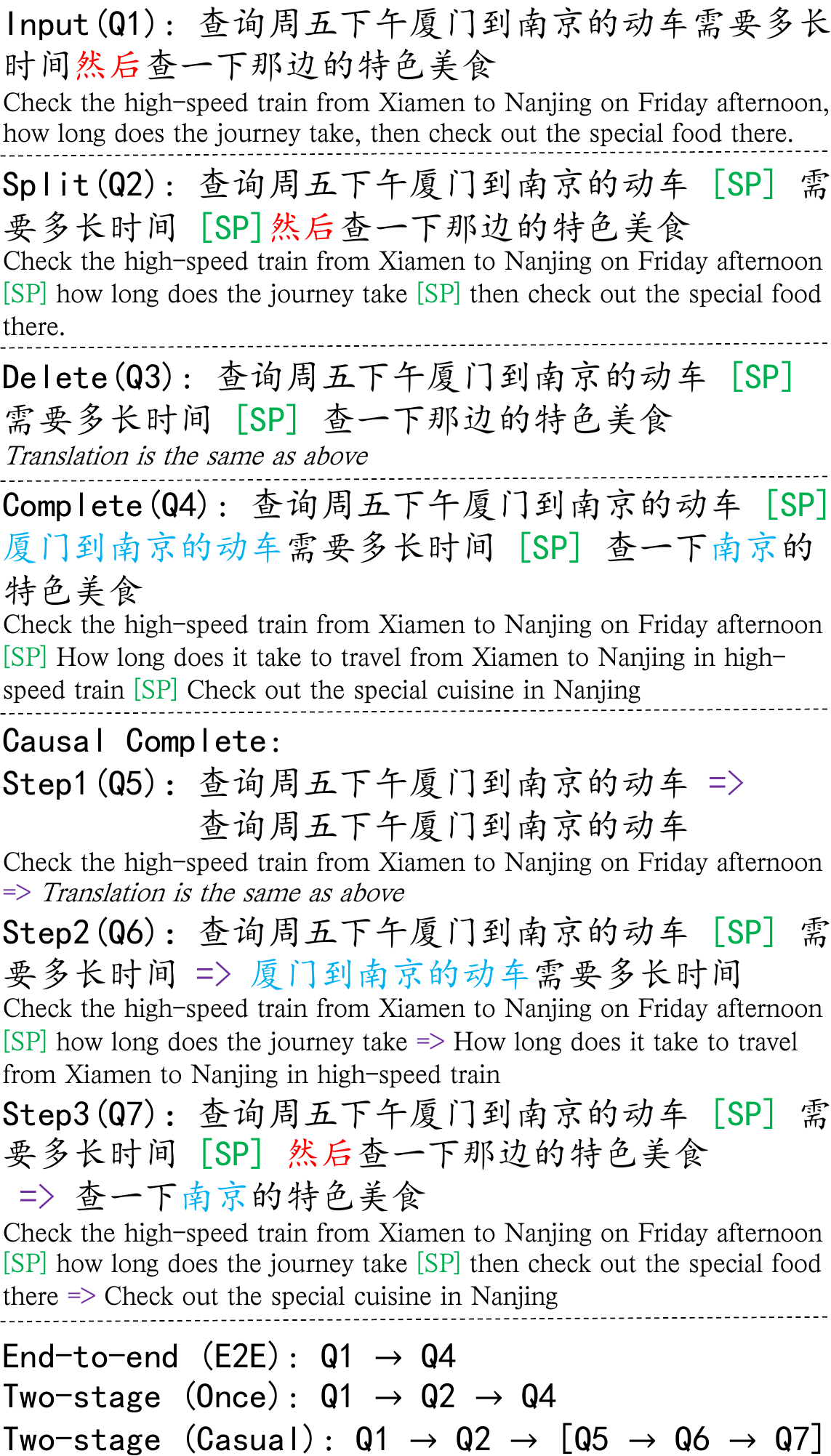}
    \caption{The overview for the actions taken to transform a multi-intent complex user query (Q1) to the executable single-intent queries (Q4). We use red, blue and green to highlight the text fillers, omitted information and query delimiters, respectively.} 
    \label{fig:method_overview}
\end{figure}

\begin{table*}[t!]
\centering
\begin{tabular}{lcccccccc}
\hline
\multirow{2}{*}{\textbf{Model}} & \multirow{2}{*}{\textbf{BLEU}} 
& \multirow{2}{*}{\textbf{METEOR}} & \multirow{2}{*}{\textbf{ROUGE}}
& \multirow{2}{*}{\textbf{SACC}} & \multicolumn{3}{c}{\textbf{Exact Match (EM)}} \\
 & & & & & Comp. & Rew. & Avg. \\
\hline
End-to-end (mT5-base) & 49.37  & 57.54 & 60.14  & 84.40 & 88.26 & 56.89 & 56.17 \\
End-to-end (mT5-large) & 57.07  & 59.83 & 66.47  & 93.50 & 93.53 & 63.04 & 63.90 \\
End-to-end (mT5-xl) & 64.37  & 62.52 & \textbf{72.71}  & 97.90 & \textbf{97.52} & 70.07 & 71.45 \\
End-to-end (mBART-large) & 62.32  & 62.24 & 71.09  & 98.60 & 97.03 & 68.11 & 69.48 \\
\hline
Two-stage (once,mT5-base) & 49.38 & 58.03 & 60.21 & 83.80 & 89.64 & 56.95 & 56.17 \\
Two-stage (casual,mT5-base) & 55.29 & 60.26 & 65.11 & 85.30 & 89.80 & 61.72 & 60.79 \\
Two-stage (once,mT5-large)& 57.86 & 60.39 & 67.03 & 94.10 & 92.54 & 63.79 & 64.53 \\
Two-stage (casual,mT5-large) & 62.09 & 62.04 & 70.68 & 94.20 & 93.53 & 67.72 & 68.24 \\
Two-stage (once,mT5-xl) & \textbf{64.37} & 62.82 & 72.50 & 98.70 & 97.04 & \textbf{70.33} & \textbf{71.78} \\
Two-stage (casual,mT5-xl) & 63.73 & \textbf{62.86} & 72.46 & \textbf{98.80} & 95.07 & 70.28 & 71.62\\
\hline
\end{tabular}
\caption{The benchmark for the baseline models (Fig \ref{fig:method_overview}). ``Comp.'' and ``Rew.'' correspond to the complete and rewritten (incomplete due to coreferences or omissions) queries. We report the median scores over 5 runs.}
\label{tab:main_res}
\end{table*}

\section{Baseline Models}
\subsection{Task Overview}
As depicted in Figure \ref{fig:method_overview}, the input (Q1) and the output (Q4) of DialogUSR have a large text overlap. The transformation from Q1 to Q4 can be viewed as several local edits that retain the main body of the input query.
We thus define several implicit actions that guide the transformation:
\textbf{1)} The \texttt{Split} action (Q1$\rightarrow$Q2)  divides the complex multi-intent query into specific single-intent query with a special token. In our implementation we use the semicolon (;) and set up a heuristic rule that puts the semicolons before the text fillers if the latter appear. 
\textbf{2)} The \texttt{Delete} action (Q2$\rightarrow$Q3)  removes the text fillers and keep the salient queries for the subsequent actions.
\textbf{3)} The \texttt{Complete} action (Q3$\rightarrow$Q4)  recovers the coreferred and omitted information in the recognized single-intent queries so that they can be effectively parsed by the existing (single-query) NLU module.  
\textbf{4)} The \texttt{Causal Complete} strategy consists of the \texttt{Split} action (Q1$\rightarrow$Q2) and several \texttt{Complete} actions that echo with the token-by-token auto-regressive text generation. The difference is that \texttt{Causal Complete} strategy in DialogUSR recovers the missing information in the incomplete user utterances with a query-by-query fashion (Q5$\rightarrow$Q6$\rightarrow$Q7).

\subsection{End-to-end Generative Models}
The most straightforward way is to train a sequence-to-sequence model to learn the transformation from the multi-intent query (Q1) to the decomposed single-intent ones (Q4) in the end-to-end fashion.
The models are trained to implicitly split the raw query (without punctuation) (Q1$\rightarrow$Q2), delete the conjunctions (Q2$\rightarrow$Q3) and recover the missing information (Q3$\rightarrow$Q4) in one single turn of generation. Specifically given the multi-intention complex query, the model is trained to output the sequence of multiple completed independent queries ``$Q_1; Q_2; ...; Q_n;$ </s>'', where ``;'', $n$,  ``</s>'' represent the query separation token, the number of queries and the end-of-sentence token, respectively. 
\subsection{Two-stage Generative Models}
\label{sec:two_stage_baseline}
In stead of performing all three actions in one single turn, 
we try to guide the transformation by a step-by-step generation \citep{moryossef-etal-2019-step,liu-etal-2019-towards-comprehensive,DBLP:conf/aaai/0001ZCS21}. 
Notably, the \texttt{Split}, \texttt{Delete} and \texttt{Complete} actions in Fig \ref{fig:method_overview} can be arbitrarily permuted throughout the generation process, e.g. firstly removing text filler then split the complete the complex query (\texttt{Delete}$\rightarrow$\texttt{Split}$\rightarrow$\texttt{Complete}). 
However we observe the performance drop if we explicitly employ a 3-step generation due to the error propagation. 
\paragraph{Two-stage model (once)} 
we resort to a two-stage procedure that firstly splits the complex query (Q1$\rightarrow$Q2) and then recovers the incomplete utterances (Q2$\rightarrow$Q4). As the \texttt{Split} action is relative easy, i.e. achieving nearly 100\% accuracy on the query separation, the error accumulations are largely mitigated. 

\paragraph{Two-stage model (casual)} 
Due to the fact that the former sub-queries would not be affected by the subsequent queries, we propose a ``causal''-style query-by-query generation (Q5$\rightarrow$Q6$\rightarrow$Q7) in which the current sub-query to be reformulated only conditions on the prior sub-query instead of seeing the bidirectional context.
Specifically, the \texttt{Causal complete} action takes place after the \texttt{Split} action. 
In the $t$-th episode of \texttt{Causal complete} action, we feed the model with incomplete queries ``$q_1; ...; q_t$'', and then train the model to generate the completed query $Q_t$. In this way, we greatly reduce the search space without the sacrifice on model performance. 
From an engineering standpoint, the proposed \texttt{Causal complete} action is a natural fit for the ``streaming'' conversational agent, i.e. simultaneous query splitting and information recovery followed by single-intent NLU while the users are eliciting multiple intents.

\section{Experiment Settings}

\paragraph{Model Setting}
We experiment with a variety of pretrained models via Hugging Face Transformers \cite{wolf-etal-2020-transformers}, including mT5 \citep{xue-etal-2021-mt5} with three different parameter scales, namely T5-base (580M), T5-large (1.2B), T5-xl (3.7B) , and  mBART-large \citep{liu-etal-2020-multilingual-denoising} with 340M parameters as the backbones for the end-to-end and two-stage models. They are all multi-lingual pre-trained models that support both Chinese and English DialogUSR. We use the Adam optimizer \citep{DBLP:journals/corr/KingmaB14} with the learning rate of $0.00003$ and train the models for maximum 9 epochs on 4-8 A100 Gpus.

\paragraph{Evaluation Metrics}
Viewing DialohUSR as a sequence generation task, i.e. concatenating the segmented single-intent queries with semicolons like Q4 in Fig \ref{fig:method_overview}, we use BLEU-4 \citep{papineni-etal-2002-bleu}, METEOR \citep{10.1007/s10590-009-9059-4}, ROUGE-L \citep{lin-2004-rouge}, which are three commonly used automatic evaluation metrics to measure the ngram similarity with the reference in the token level. We also propose two new sentence-level metrics, namely Split Accuracy (SACC) and Exact Match (EM) to evaluate the model performance for DialogUSR. 
Specifically SACC measures the ratio of correct query splitting. We consider a multi-query to be correctly separated if the models split it into exactly the same number of queries as the reference:
$$
\text{SACC} = \frac{1}{n} \sum_{1 \le i \le n} {\mathbb{I}_{\text{len}(Q_{pred}^{(i)}) = \text{len}(Q_{ref}^{(i)})} },
$$
where $n$ is the number of instances, $\mathbb{I}$ is the indicator function, $Q_{pred}^{(i)}$ and $Q_{ref}^{(i)}$ are the $i$-th predicted and reference query list.
As for EM, we consider it correct if the predicted query is exactly the same as the reference one:
$$
\text{EM} = \frac{\sum_i \sum_j \mathbb{I}_{Q_{pred\_j}^{(i)} = Q_{ref\_j}^{i}} }{\sum_{1 \le i \le n} \text{len}(Q_{ref}^{(i)})},
$$
where $Q_{pred\_j}^{(i)}$ and $Q_{ref\_j}^{(i)}$ represent the $j$-th predicted and reference query of the $i$-th instance. We calculate Exact Match in three different situations: EM-Complete, where we only consider the queries that does not need further modification (Q5 in Fig \ref{fig:method_overview}); EM-Rewritten, where delete or complete actions are needed (Q6, Q7 in Fig \ref{fig:method_overview}); and finally EM-Average, in which we consider all the queries.

\begin{table}[t!]
\centering
\begin{tabular}{lllllll}
\hline
\textbf{Combination}  & \textbf{BLEU} & \textbf{SACC} & \textbf{EM}\\
\hline
\texttt{DE} $\rightarrow$ (\texttt{SP}+\texttt{CP}) & 46.08  & 77.00 & 51.25\\
(\texttt{DE}+\texttt{CP}) $\rightarrow$ \texttt{SP} & 34.23  & 70.10 & 47.57\\
(\texttt{SP}+\texttt{DE}) $\rightarrow$ \texttt{CP} & 47.60  & 82.00 & 54.15\\
(\texttt{SP}+\texttt{CP}) $\rightarrow$ \texttt{DE} & 45.67  & 80.20 & 52.42\\
\texttt{CP} $\rightarrow$ (\texttt{SP}+\texttt{DE})   & 45.66  & 78.80 & 52.28 \\
\texttt{SP} $\rightarrow$ \texttt{DE} $\rightarrow$ \texttt{CP}     & 47.66  & 83.40 & 54.48   \\
\hline
\texttt{SP} $\rightarrow$ (\texttt{DE}+\texttt{CP}) & \textbf{49.37} & \textbf{84.40} & \textbf{56.17} \\
\hline
\end{tabular}
\caption{The exploration on the most effective action combination for the two-stage (once) model using the mT5-base models. \texttt{SP}, \texttt{DE}, \texttt{CP} are the abbreviations of the \texttt{Split}, \texttt{Delete} and \texttt{Complete} actions in Fig \ref{fig:method_overview}.} 
\label{tab:two_stage_combine}
\end{table}

\begin{table}[t!]
\centering
\begin{tabular}{lcccc}
\hline
\multirow{2}{*}{\textbf{Model}}  
& \multicolumn{2}{c}{\textbf{MixSNIPS}} & \multicolumn{2}{c}{\textbf{MixATIS}} \\
 & BLEU & EM & BLEU & EM \\
\hline
T5-base & 99.46 & 95.13 & 96.94 & 74.88 \\
T5-large & 99.60 & 97.64 & 98.52 & 88.77 \\
T5-xl & \textbf{99.62}  & \textbf{98.14} & \textbf{99.87} & \textbf{98.55} \\
\hline
\end{tabular}
\caption{End-to-end model performance on the MixSNIPS and MixATIS datasets.}
\label{tab:mix_snips_atis}
\end{table}


\begin{figure}[t!]
    \centering
    \includegraphics[width=1.0\linewidth]{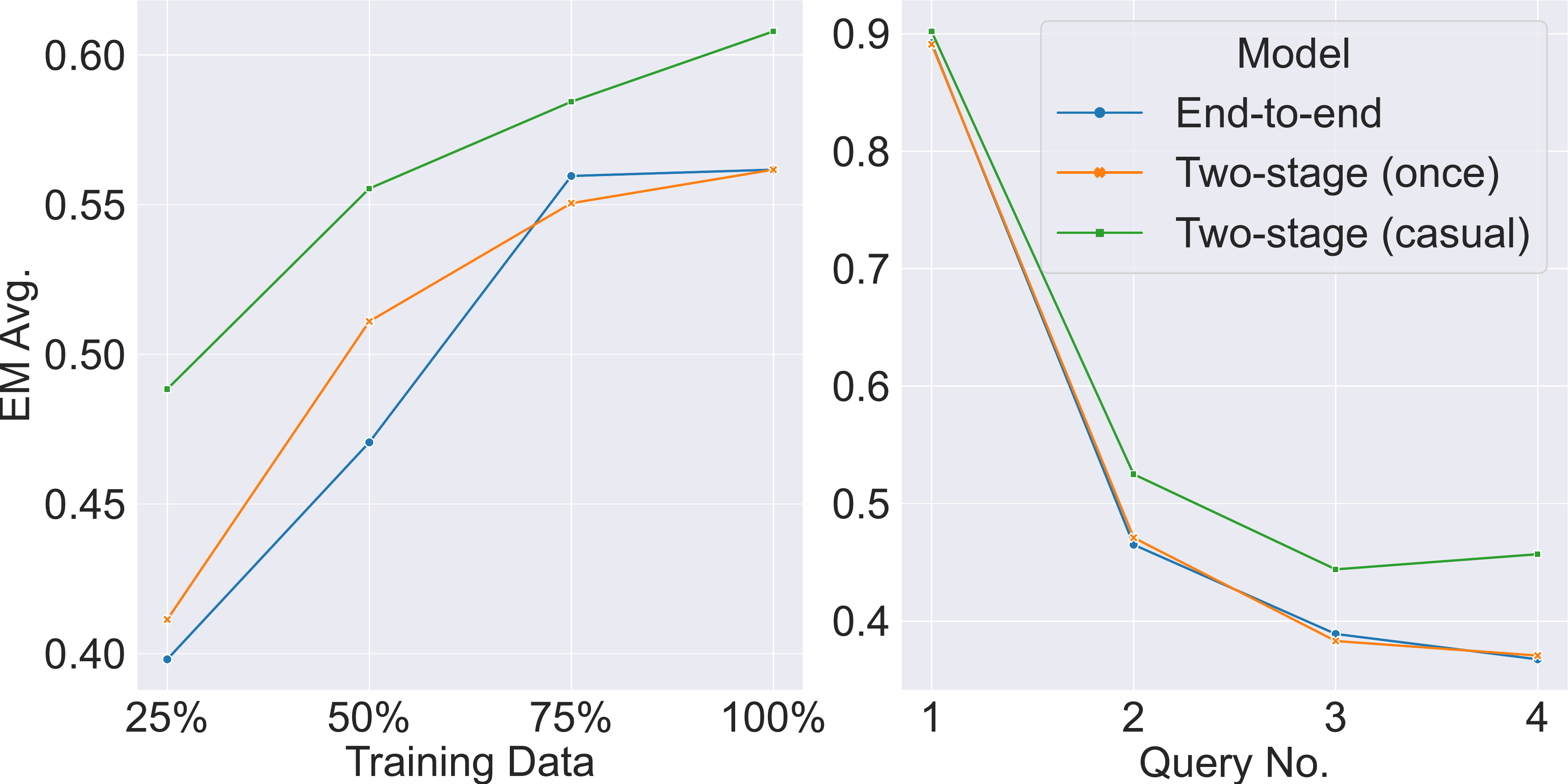}    
    \caption{ The model performance (mT5-base) of different training data scale (left) and sub-queries (right).
    }
    \label{fig:small_data}
\end{figure}

\begin{figure}[t!]
    \centering
    \includegraphics[width=1.0\linewidth]{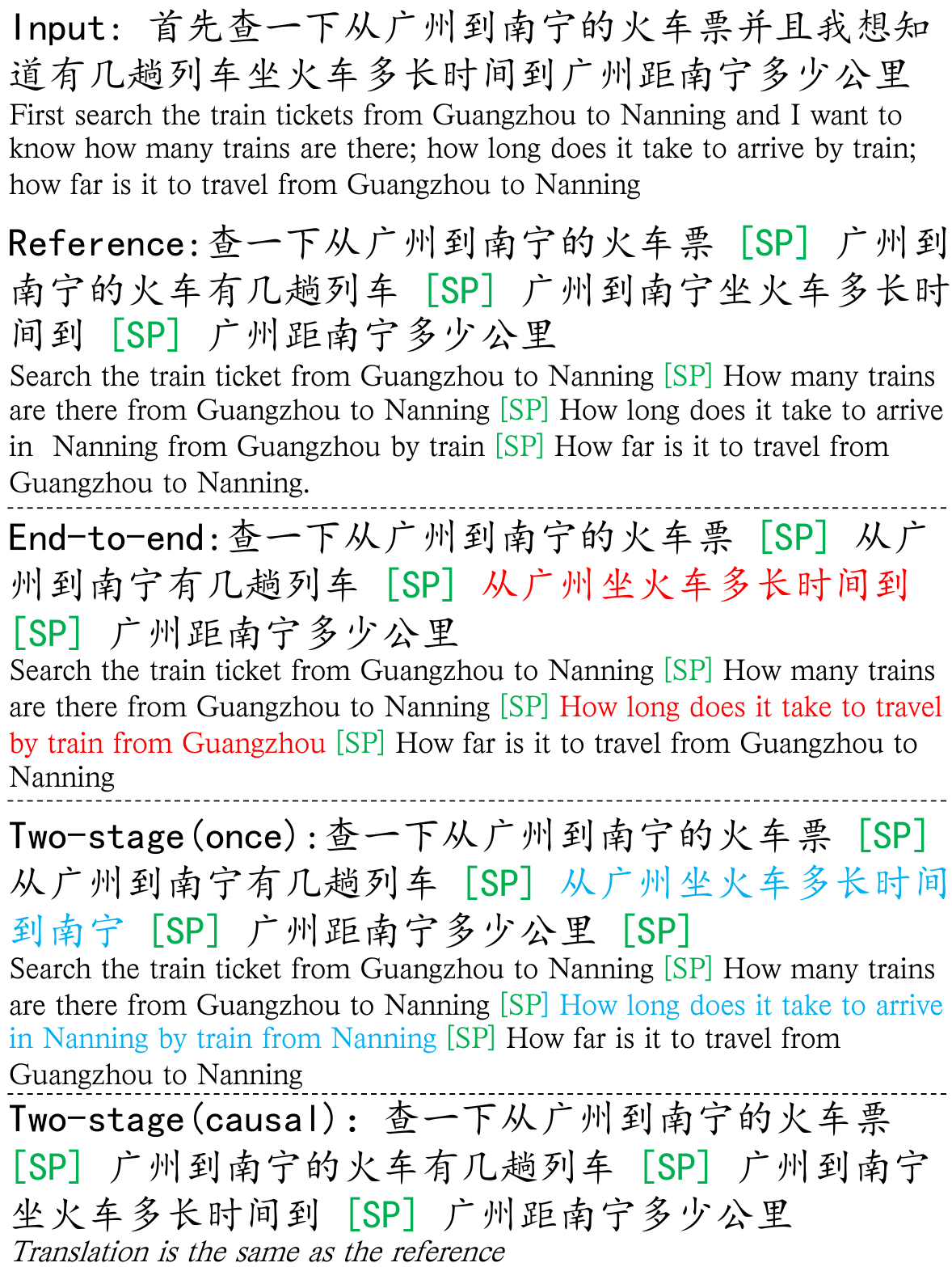}    
    \caption{ The demonstration of generated outputs for different baseline models. The query marked in red is wrong due to the missing destination of the train, while the query marked in blue is a paraphrase of the the reference.
    }
    \label{fig:case_show}
\end{figure}

\section{Analysis and Discussions}
\paragraph{Baseline performance}
Table \ref{tab:main_res} shows the performance of the baseline models on DialogUSR. For both end-to-end and two-stage generative baselines, enlarging the model parameters of mT5 models leads to a considerable performance gain, which indicates that powerful pretrained models with larger capacity are important in learning query transformation in DialogUSR.
In terms of the comparisons between end-to-end and two variants of two-stage models, we observe that for mT5-base and mT5-large, the causal-style two-stage model is the clear winner among the three models, which shows that the query-by-query transformation (Q5$\rightarrow$Q6$\rightarrow$Q7 in Fig \ref{fig:method_overview}) is the most effective way to recover the missing information while reformulating the queries. For mT5-xl, the performance gap between two-stage and end-to-end baselines is largely reduced, indicating powerful trained models may close the gap between different baselines. 

We also report the model performance on the existing multi-intent detection datasets, namely MixSNIPS and MixATIS.
As mentioned in Sec \ref{sec:data_analysis}, both of them are created by inserting specific conjunctions between two complete single-intent queries from the SNIPS \citep{coucke2018snips} or ATIS \citep{hemphill-etal-1990-atis} datasets, without any coreference or omission phenomenon. In other words, both of them can be effectively solved with an end-to-end model using the \texttt{Delete} and \texttt{Split} actions. The large performance gap of the same model on the MixATIS/ MixSNIPS and the proposed DialogUSR verifies that the multi-intent query splitting and reformulation task is far from solved.

\paragraph{Findings in the different action combinations}
As elaborated in Sec \ref{sec:two_stage_baseline} and Fig \ref{fig:method_overview}, the \texttt{Split}, \texttt{Delete} and \texttt{Complete} actions can be permuted during the generation. We thus try to find the most effective action combination for the two-stage (once) model as shown in Table \ref{tab:two_stage_combine}.
We find that 
\textbf{1)} The 3-stage models\footnote{We try different actions permutations on the 3-stage models and put the most effective combination in Table \ref{tab:two_stage_combine}.} (\texttt{SP}$\rightarrow$\texttt{DE}$\rightarrow$ \texttt{CP})
are not necessary in the multi-stage generation compared with its two-stage variants (\texttt{SP} $\rightarrow$ (\texttt{DE}+\texttt{CP})) because of the risk of error propagation (performance drop) and larger computational overhead.
\textbf{2)} The \texttt{Split} action should be placed in the first stage, as placing it in the second stage exhibit large performance drop, e.g. \texttt{SP} $\rightarrow$  (\texttt{DE}+\texttt{CP}) and (\texttt{DE}+\texttt{CP}) $\rightarrow$ \texttt{SP}.
Presumably this is because the query splitting transformation may not be robust to the potentially ill-formed rewritten queries due to the lack of exposure to the noisy training data.
\textbf{3)} The \texttt{Delete} and \texttt{Complete} actions should be merged and placed in the second stage of generation. These two actions together can be viewed as a rewriting operation that deletes the conjunctions and recovers the missing information.

\paragraph{Detailed analysis on model outputs}
As the DialogUSR is actually a domain-agnostic query rewriting task, we investigate the performances of the baseline models with different training data scale in Fig \ref{fig:small_data} (left). With less training data, we observe a clear boost while employing the two-stage models.
Fig \ref{fig:small_data} (right) shows the model performance while generating the sub-queries in different positions, e.g. Q5, Q6, Q7 in Fig \ref{fig:method_overview}) correspond to the first, second and third queries while splitting the multi-intent complex query.
We observe a large performance drop while comparing the first query and the subsequent queries, because in real-world scenarios most users would not include coreferences or omissions in the query, which make it much easier to split and complete the first sub-query.

We also provide a case study for the generated outputs from different baseline models in Fig \ref{fig:case_show}. Both the models trained with the two-stage strategy produce correct and executable single queries, while the end-to-end model misses the destination information in the third query, which would end up with the false parsing results in the downstream NLU modules of conversational agents.

\section{Related Work}
\paragraph{Incomplete Utterance Restoration}
To convert multi-turn incomplete dialogue into multiple single-turn complete utterance, two major paradigms are available currently. One straight-forward way is to consider it as a sequence-to-sequence problem, using models including RNN \citep{pan-etal-2019-improving, elgohary-etal-2019-unpack}, Trans-PG+BERT \citep{hao-etal-2021-rast} and T5 with importance token selection \citep{DBLP:journals/corr/abs-2204-03958}. And since the source and target utterances are highly overlapped, another approach is to edit rather than generate from scratch, specifying the operation by sequence tagging. \citet{pan-etal-2019-improving} proposed Pick-and-Combine model, while \citet{liu-etal-2020-incomplete} introduced Rewritten U-shaped Network which imitates semantic segmentation by predicting the word-level edit matrix, and with similarity \citet{Huang_Li_Zou_Zhang_2021} used a semi auto-regressive generator. Later, \citet{hao-etal-2021-rast} proposed RUST to address the robustness issue and \citet{jin2022hierarchical} proposed hierarchical context tagging to achieve higher phrase coverage.
\paragraph{Multi-intent detection}
Spoken language understanding (SLU) which consists of intent detection and slot filling is the core in  spoken dialogue systems\citep{tur2011spoken}. Intent detection mainly aims to classify a given utterance with its intents from user inputs. Considering this strong correlation between the two tasks, some joint models are proposed based on the multi-task learning framework. \citep{zhang2016joint,goo-etal-2018-slot,qin-etal-2019-stack,7078572,li-etal-2018-self}. \citet{li-etal-2018-self} proposed the gate mechanism to explore incorporating the intent information for slot filling. Convolutional-LSTM and capsule network have been proposed to solve the problem \citep{xia-etal-2018-zero}.  \citet{gangadharaiah-narayanaswamy-2019-joint} shows that 52\% utterances are multi-intent in the Amazon internal dataset which indicate that in real world scenario, however, users often input utterance containing multi-intent. Therefore, \citet{Rychalska2018MultiIntentHN} first adopted hierarchical structures to identify multiple user intents.  \citet{qin-etal-2020-agif} associate multi-intent detection with slots filling via graph attention network.

\citet{larson2022survey} offers a thorough overview on the existing multi-intent detection datasets. Except from MixATIS and MixSNIPS datasets,
TOP \citep{gupta-etal-2018-semantic-parsing} contains multi-intent queries annotated in a hierarchical manner which dramatically improves the expressive power while DialogUSR contains queries and rewriting queries which can bridge the single-intent dection and multi-intent detection and also decoupling the query intent detection section and multi-intent query separation section.
NLU++ \citep{casanueva-etal-2022-nlu} has been collected, filtered and carefully annotated by dialogue NLU experts while DialogUSR queries are created by human annotators and aggregated by rules and evaluated by model which lead to a lower cost of data annotation than NLU++.

\section{Conclusion}
We propose DialogUSR, a dialog utterance splitting and reformulation task and corresponding dataset, for multi-intent detection in the conversational agents.
The model trained on DialogUSR can serve as a domain-agnostic and plug-in module for the existing product chatbots with minial efforts.
The proposed dataset contains 11.6k high quality instances that cover 23 domains with a multi-step annotation process.
We propose multiple action-based generative baselines to benchmark the dataset and analyze their pros and cons through a series of investigations.

\section*{Limitations}
The proposed DialogUSR focuses on a single task for the research community and lacks of implementation details in the product conversational agents.
The approaches on how the proposed DialogUSR interacts with other modules, e.g. dialog manager, ranking module for candidate NLU parsing results, remains an interesting and important research area.
We position our work in the line of researches which enhances advanced conversational AI (i.e. multi-turn or multi-intent) by \emph{query rewriting}, and leave multi-intent slot-filling entity annotation to the further work.


\section*{Acknowledgement}

We thank all the anonymous reviewers for their insightful feedback. This paper is supported by the National Key Research and Development Program of China 2020AAA0106700 and the National Natural Science Foundation of China (NSFC) project U19A2065.

\bibliography{emnlp2022}
\bibliographystyle{acl_natbib}

\clearpage
\newpage

\appendix

\section{Implementation Detail}
We run the experiments with Huggingface Transformers library on 4 Nvidia A100 GPU, with the training batch size of 96. For experiments on full training set, we set warm up step as 50. For beam search of seq2seq model, the beam size is 4. We train and test our models with 8 A100 GPUs.

\begin{table}[ht]
\centering
\small
\begin{tabular}{lcc}
\hline
Model-Size &  Inference (ms) & Train (s) \\
\hline
End-to-end, base & 71.94  & 289 \\
End-to-end, large & 114.95  & 450 \\
End-to-end, xl & 118.59  & 712 \\
Two-stage(once), base & 156.85  & 570 \\
Two-stage(once), large & 201.11  & 875 \\
Two-stage(once), xl & 211.26  & 1392 \\
Two-stage(casual), base & 173.90  & 1118 \\
Two-stage(casual), large & 253.06  & 1911 \\
Two-stage(casual), xl & 277.07  & 2859 \\
\hline
\end{tabular}
\label{table:inference_training_time}
\caption{Model Inference Speed and Training Time}
\end{table}

\begin{table}[t!]
\centering
\resizebox{\linewidth}{!}{
\begin{tabular}{lllll}
\hline
       & Conj0\_Prob & Conj1\_Prob & Conj2\_Prob & Conj3\_Prob  \\
\hline
None   & 50\%              & 50\%              & 50\%              & 50\%               \\
\begin{CJK*}{UTF8}{gbsn}先\end{CJK*}      & 25\%              & 0\%                 & 0\%                 & 0\%                  \\
\begin{CJK*}{UTF8}{gbsn}首先\end{CJK*}    & 25\%              & 0\%                 & 0\%                 & 0\%                  \\
\begin{CJK*}{UTF8}{gbsn}然后\end{CJK*}     & 0\%               & 10\%              & 10\%              & 10\%               \\
\begin{CJK*}{UTF8}{gbsn}还有\end{CJK*}     & 0\%               & 2.50\%            & 2.50\%            & 2.35\%             \\
\begin{CJK*}{UTF8}{gbsn}我还想知道\end{CJK*}  & 0\%               & 2.50\%            & 2.50\%            & 2.35\%             \\
\begin{CJK*}{UTF8}{gbsn}另外我想知道\end{CJK*} & 0\%               & 2.50\%            & 2.50\%            & 2.35\%             \\
\begin{CJK*}{UTF8}{gbsn}再一个就是\end{CJK*}  & 0\%               & 2.50\%            & 2.50\%            & 2.35\%             \\
\begin{CJK*}{UTF8}{gbsn}以及\end{CJK*}     & 0\%               & 2.50\%            & 2.50\%            & 2.35\%             \\
\begin{CJK*}{UTF8}{gbsn}和\end{CJK*}      & 0\%               & 2.50\%            & 2.50\%            & 2.35\%             \\
\begin{CJK*}{UTF8}{gbsn}还要\end{CJK*}     & 0\%               & 2.50\%            & 2.50\%            & 2.35\%             \\
\begin{CJK*}{UTF8}{gbsn}并且\end{CJK*}     & 0\%               & 2.50\%            & 2.50\%            & 2.35\%             \\
\begin{CJK*}{UTF8}{gbsn}再然后\end{CJK*}    & 0\%               & 2.50\%            & 2.50\%            & 2.35\%             \\
\begin{CJK*}{UTF8}{gbsn}另外\end{CJK*}     & 0\%               & 2.50\%            & 2.50\%            & 2.35\%             \\
\begin{CJK*}{UTF8}{gbsn}其次\end{CJK*}     & 0\%               & 2.50\%            & 2.50\%            & 2.35\%             \\
\begin{CJK*}{UTF8}{gbsn}同时\end{CJK*}     & 0\%               & 2.50\%            & 2.50\%            & 2.35\%             \\
\begin{CJK*}{UTF8}{gbsn}除了这个还有\end{CJK*} & 0\%               & 2.50\%            & 2.50\%            & 2.35\%             \\
\begin{CJK*}{UTF8}{gbsn}接着\end{CJK*}     & 0\%               & 2.50\%            & 2.50\%            & 2.35\%             \\
\begin{CJK*}{UTF8}{gbsn}紧接着\end{CJK*}    & 0\%               & 2.50\%            & 2.50\%            & 2.35\%             \\
\begin{CJK*}{UTF8}{gbsn}接下来\end{CJK*}    & 0\%               & 2.50\%            & 2.50\%            & 2.35\%             \\
\begin{CJK*}{UTF8}{gbsn}最后\end{CJK*}    & 0\%               & 0.00\%            & 0.00\%            & 2.35\% \\
\hline  
\end{tabular}
}
\caption{Conjunction probability distribution in four queries cases. }
\label{conjection-4-query}
\end{table}

\begin{table}[t!]
\centering
\resizebox{\linewidth}{!}{
\begin{tabular}{llll}
\hline 
       & Conj0\_Prob & Conj1\_Prob & Conj2\_Prob  \\
       \hline 
None   & 50\%              & 50\%              & 50\%               \\
\begin{CJK*}{UTF8}{gbsn}先\end{CJK*}      & 25\%              & 0\%                 & 0\%                  \\
\begin{CJK*}{UTF8}{gbsn}首先\end{CJK*}    & 25\%              & 0\%                 & 0\%                  \\
\begin{CJK*}{UTF8}{gbsn}然后\end{CJK*}    & 0\%               & 10\%              & 10\%               \\
\begin{CJK*}{UTF8}{gbsn}还有\end{CJK*}    & 0\%               & 2.50\%            & 2.35\%             \\
\begin{CJK*}{UTF8}{gbsn}我还想知道\end{CJK*}  & 0\%               & 2.50\%            & 2.35\%             \\
\begin{CJK*}{UTF8}{gbsn}另外我想知道\end{CJK*} & 0\%               & 2.50\%            & 2.35\%             \\
\begin{CJK*}{UTF8}{gbsn}再一个就是\end{CJK*}  & 0\%               & 2.50\%            & 2.35\%             \\
\begin{CJK*}{UTF8}{gbsn}以及\end{CJK*}     & 0\%               & 2.50\%            & 2.35\%             \\
\begin{CJK*}{UTF8}{gbsn}和\end{CJK*}     & 0\%               & 2.50\%            & 2.35\%             \\
\begin{CJK*}{UTF8}{gbsn}还要\end{CJK*}     & 0\%               & 2.50\%            & 2.35\%             \\
\begin{CJK*}{UTF8}{gbsn}并且\end{CJK*}    & 0\%               & 2.50\%            & 2.35\%             \\
\begin{CJK*}{UTF8}{gbsn}再然后\end{CJK*}    & 0\%               & 2.50\%            & 2.35\%             \\
\begin{CJK*}{UTF8}{gbsn}另外\end{CJK*}    & 0\%               & 2.50\%            & 2.35\%             \\
\begin{CJK*}{UTF8}{gbsn}其次\end{CJK*}     & 0\%               & 2.50\%            & 2.35\%             \\
\begin{CJK*}{UTF8}{gbsn}同时\end{CJK*}     & 0\%               & 2.50\%            & 2.35\%             \\
\begin{CJK*}{UTF8}{gbsn}除了这个还有\end{CJK*} & 0\%               & 2.50\%            & 2.35\%             \\
\begin{CJK*}{UTF8}{gbsn}接着\end{CJK*}     & 0\%               & 2.50\%            & 2.35\%             \\
\begin{CJK*}{UTF8}{gbsn}紧接着\end{CJK*}     & 0\%               & 2.50\%            & 2.35\%             \\
\begin{CJK*}{UTF8}{gbsn}接下来\end{CJK*}    & 0\%               & 2.50\%            & 2.35\%             \\
\begin{CJK*}{UTF8}{gbsn}最后\end{CJK*}     & 0\%               & 0.00\%            & 2.35\%            \\
\hline 
\end{tabular}
}
\caption{Conjunction probability distribution in three queries cases.}
\label{conjection-3-query}
\end{table}

\begin{table*}[h]
\small
\centering
\begin{tabular}{lp{12.5cm}}
\hline 
\\ [-0.9ex] 
\textbf{Input} & \begin{CJK*}{UTF8}{gbsn}昭通到南昌的汽车最早几点发车走高速吗另外我想知道自驾要多久到\end{CJK*} \\
 & The bus from Zhaotong to Nanchang; what time does it leave at the earliest; does it take the highway; also I would like to know how long it will take to get there by self-driving \\ 
\hline
\\ [-0.9ex] 
\textbf{End-to-end} & \begin{CJK*}{UTF8}{gbsn}昭通到南昌的汽车 [SP] 昭通到南昌的汽车最早几点发车 [SP] 昭通到南昌的汽车走高速吗 [SP] 昭通到南昌自驾要多久到 \end{CJK*} \\ 
 & The bus from Zhaotong to Nanchang [SP] What time does the bus from Zhaotong to Nanchang leave at the earliest [SP] Does the bus from Zhaotong to Nanchang take the highway [SP] How long it will take to get there from Zhaotong to Nanchang by self-driving \\
\textbf{Two-stage (Once)} & \begin{CJK*}{UTF8}{gbsn}昭通到南昌的汽车 [SP] 昭通到南昌的汽车最早几点发车 [SP] 昭通到南昌的汽车走高速吗 [SP]  自驾昭通到南昌要多久到 \end{CJK*} \\
 & The bus from Zhaotong to Nanchang [SP] What time does the bus from Zhaotong to Nanchang leave at the earliest [SP] Does the bus from Zhaotong to Nanchang take the highway [SP] How long it will take to get there by self-driving from Zhaotong to Nanchang   \\
\textbf{Two-stage (causal)} & \begin{CJK*}{UTF8}{gbsn}昭通到南昌的汽车 [SP] 昭通到南昌的汽车最早几点发车 [SP] 昭通到南昌的汽车走高速吗 [SP]  自驾昭通到南昌要多久到 \end{CJK*}  \\
 & The bus from Zhaotong to Nanchang [SP] What time does the bus from Zhaotong to Nanchang leave at the earliest [SP] Does the bus from Zhaotong to Nanchang take the highway [SP] How long it will take to get there by self-driving from Zhaotong to Nanchang \\

\hline
\\
\hline 
\\ [-0.9ex] 
\textbf{Input} & \begin{CJK*}{UTF8}{gbsn}推荐一部动作电影我要看三十集左右的电视剧然后找一部90年代的动作电影\end{CJK*} \\
 & Recommend an action movie; I want to watch a TV series with about 30 episodes; and then look for an action movie in the 90s \\ 
\hline
\\ [-0.9ex] 
\textbf{End-to-end} & \begin{CJK*}{UTF8}{gbsn}推荐一部动作电影 [SP] 我要看三十集左右的动作电视剧 [SP] 找一部90年代的动作电影\end{CJK*} \\ 
 & Recommend an action movie [SP] I want to watch an action TV series with around thirty episodes [SP] Looking for an action movie from the 90s \\
\textbf{Two-stage (Once)} & \begin{CJK*}{UTF8}{gbsn}推荐一部动作电影 [SP] 我要看三十集左右的电视剧 [SP] 找一部90年代的动作电影 \end{CJK*} \\
 & Recommend an action movie [SP] I want to watch a TV series with around thirty episodes [SP] Looking for an action movie from the 90s   \\
\textbf{Two-stage (causal)} & \begin{CJK*}{UTF8}{gbsn}推荐一部动作电影 [SP] 我要看三十集左右的动作电视剧 [SP] 找一部90年代三十集左右的动作电影\end{CJK*}  \\
 & Recommend an action movie [SP] I want to watch an action TV series with around thirty episodes [SP] Looking for an action movie from the 90s with around thirty episodes \\
\hline
\\
\hline 
\\ [-0.9ex] 
\textbf{Input} & \begin{CJK*}{UTF8}{gbsn}首先查下下周三杭州是什么天气另外我想知道下周三有没有去杭州的航班我还想知道适合穿什么衣服\end{CJK*} \\
 & First, I want to check what weather will be in Hangzhou next Wednesday; in addition, I want to know if there is a flight to Hangzhou next Wednesday; I also want to know what clothes to wear. \\ 
\hline
\\ [-0.9ex] 
\textbf{End-to-end} & \begin{CJK*}{UTF8}{gbsn}查下下周三杭州是什么天气  [SP] 下周三有没有去杭州的航班 [SP] 下周三杭州的天气适合穿什么衣服 \end{CJK*} \\ 
 & Check what weather will be in Hangzhou next Wednesday [SP] Is there a flight to Hangzhou next Wednesday [SP] What clothes should I wear for the weather in Hangzhou next Wednesday \\
\textbf{Two-stage (Once)} & \begin{CJK*}{UTF8}{gbsn} 查下下周三杭州是什么天气 [SP] 下周三有没有去杭州的航班 [SP] 下周三杭州适合穿什么衣服 \end{CJK*} \\
 & Check what weather will be in Hangzhou next Wednesday [SP] Is there a flight to Hangzhou next Wednesday [SP] What clothes should I wear in Hangzhou next Wednesday \\
\textbf{Two-stage (causal)} & \begin{CJK*}{UTF8}{gbsn}查下下周三杭州是什么天气 [SP] 下周三有没有去杭州的航班 [SP] 下周三杭州适合穿什么衣服 \end{CJK*}  \\
 & Check what weather will be in Hangzhou next Wednesday [SP] Is there a flight to Hangzhou next Wednesday [SP] What clothes should I wear in Hangzhou next Wednesday \\
 
 \hline
\\
\hline 
\\ [-0.9ex] 
\textbf{Input} & \begin{CJK*}{UTF8}{gbsn}推荐一款越野的三厢车有没有四驱的我想要高配车\end{CJK*} \\
 & Recommend an off-road sedan; is there any four-wheel drive; I want a high-end car \\ 
\hline
\\ [-0.9ex] 
\textbf{End-to-end} & \begin{CJK*}{UTF8}{gbsn}推荐一款越野的三厢车 [SP] 有没有四驱的越野的三厢车 [SP] 我想要高配的越野的三厢车\end{CJK*} \\ 
 & Recommend an off-road sedan [SP] Is there a four-wheel drive off-road sedan [SP] I want a high-profile off-road sedan\\
\textbf{Two-stage (Once)} & \begin{CJK*}{UTF8}{gbsn}推荐一款越野的三厢车 [SP] 有没有四驱的越野三厢车 [SP] 我想要高配的四驱越野三厢车; \end{CJK*} \\
 & Recommend an off-road sedan [SP] Is there a four-wheel drive off-road sedan [SP] I want a high-end four-wheel drive off-road sedan   \\
\textbf{Two-stage (causal)} & \begin{CJK*}{UTF8}{gbsn}推荐一款越野的三厢车 [SP] 有没有四驱的越野的三厢车 [SP] 我想要高配的四驱的越野三厢车 [SP]  \end{CJK*}  \\
 & Recommend an off-road sedan [SP] Is there a four-wheel drive off-road sedan [SP] I want a high-end four-wheel drive off-road sedan\\

\hline
\end{tabular}
\label{example:2}
\caption{Examples of model rewriting queries}
\end{table*}

\section{Query Aggregation Detail}
In Table \ref{conjection-4-query}, we provide conjunction probability distribution when we have four queries need to be aggregated. Conjunction0 is placed at the head of consecutive query1. Conjunction1, Conjunction2 and Conjunction3 is placed at the tail of consecutive query1, query2 and query3 respectively. As described in Table \ref{conjection-4-query}, Conjunction0 have 50\% chance to be empty and 25\% probability to be \begin{CJK*}{UTF8}{gbsn}“先”\end{CJK*}(first) and another 25\% chance to be \begin{CJK*}{UTF8}{gbsn}“首先”\end{CJK*}(first of all). Similarly, Conjunction1 is placed at the middle of query1 and query2 with the probability described in the table and so on. Table \ref{conjection-3-query} shows the probability distribution of conjunctions when three consecutive queries that need to be aggregated. We generate 10 candidate multi-intent queries by joining consecutive queries with conjunctions described in Table \ref{conjection-4-query} and Table \ref{conjection-3-query}. After query aggregation, we calculate the perplexity of ten candidate multi-intent queries and select the most fluent sentence as multi-intent query in DialogUSR.


\section{DialogUSR Cases In All Domains}
\begin{figure*}[t]
    \centering
    \includegraphics[width=1.0\linewidth]{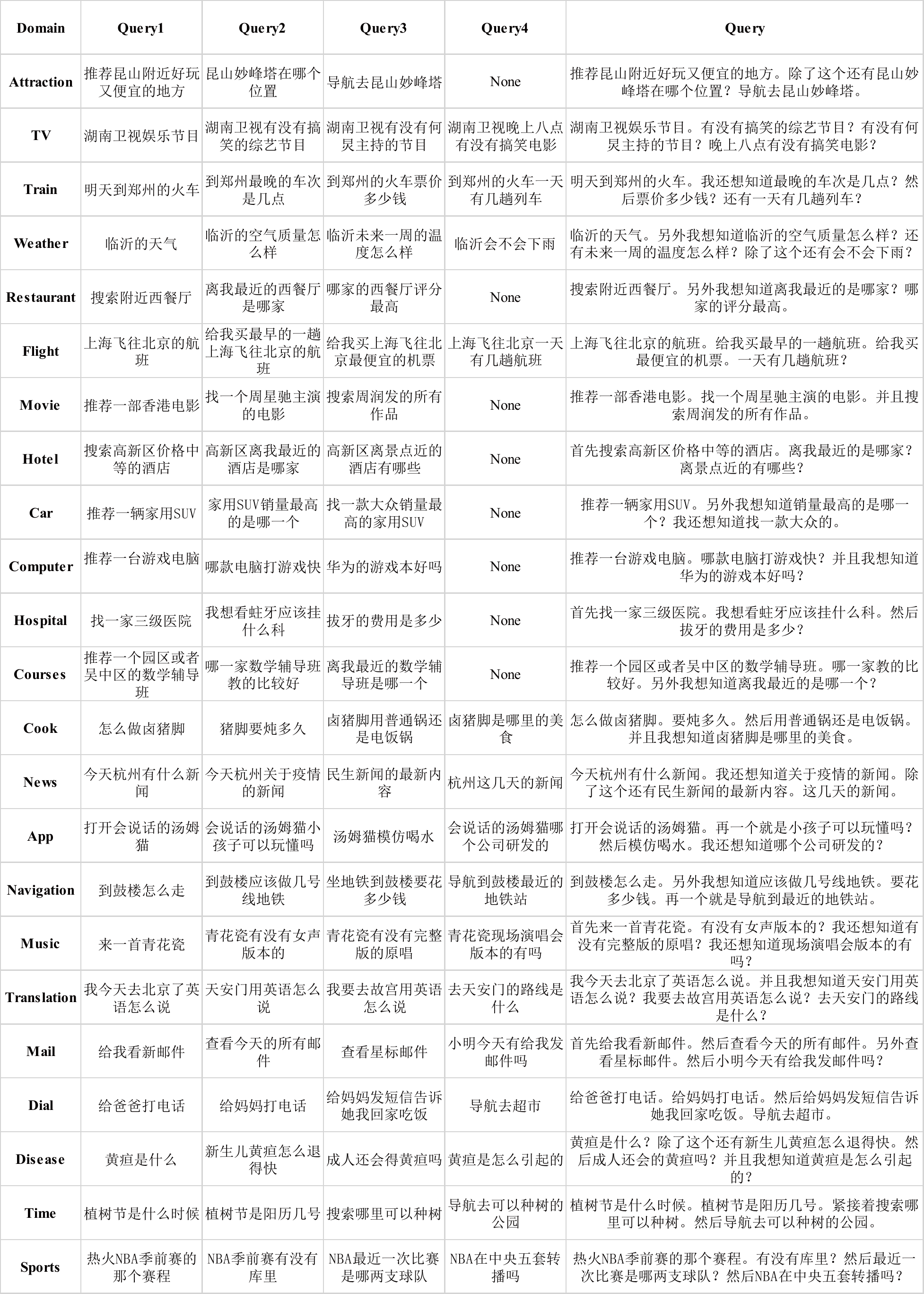}    
    \caption{DialogUSR dataset instances in all domains. Punctuations are added in the last column for better readability. }
    \label{fig:chinese domain case}
\end{figure*}

In Figure \ref{fig:chinese domain case}, for every twenty three domains, we  respectively provide one case to show our dataset. All twenty three domains in DialogUSR are listed in Figure \ref{fig:chinese domain case} including Attraction, TV, Railway, Weather, Restaurant, Flight, Movie, Hotel, Car, Hospital, Courses, Cook, News, App, Navigation, Music, Translation, Mail, Dial, Disease, Time, Sports. Query indicates the multi-intent query in DialogUSR and Query1 to Query4 represent the single-intent queries in DialogUSR.

\begin{figure*}[t]
    \centering
    \includegraphics[width=1.0\linewidth]{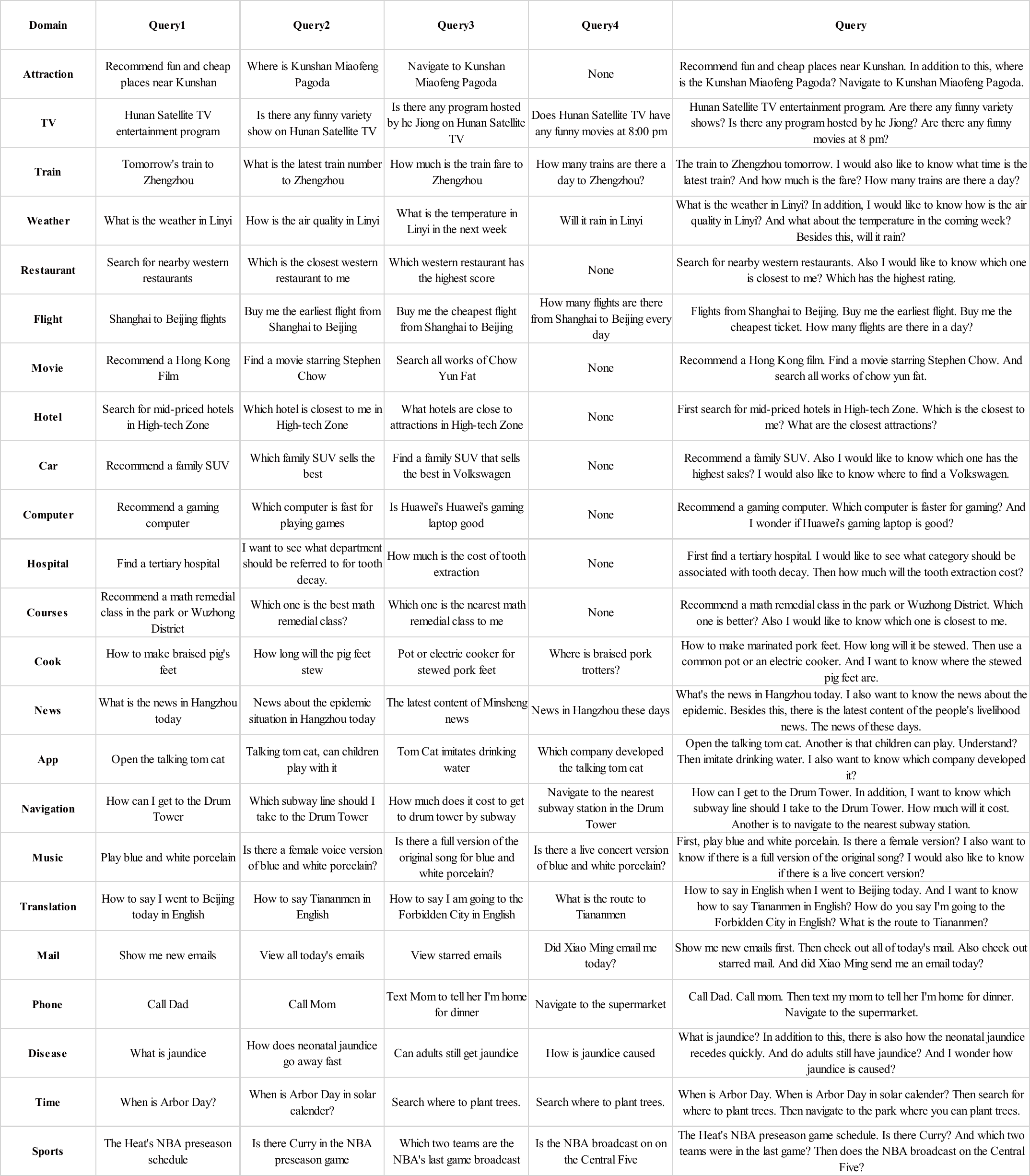}    
    \caption{English version of DialogUSR dataset instances in all domains.}
    \label{fig:english domain case}
\end{figure*}

As mentioned in Follow-up Query Creation section, we observe that 37.3\% multi-intent queries involve topic switching and this phenomenon can be found in case of Translation, Time, Phone Courses etc. In Translation case, query1 to query3 is about translation while query4 What is the route to Tiananmen(\begin{CJK*}{UTF8}{gbsn}去天安门的路线是什么\end{CJK*}) is about Navigation. As shown in Figure \ref{fig:type_statistics}, a large amount of sub-queries in multi-intent query is missing information therefore they need to be rewritten by human annotators. This situation can be easily found in many cases, for example, TV case, Railway case, Weather case, Restaurant case etc. For example, in Railway case, the sub-query I would also like to know what time is the
latest train? (\begin{CJK*}{UTF8}{gbsn}我还想知道最晚的车次是几点？\end{CJK*}) lack the key information and human annotator rewirte the sub-query as What is the latest train number to Zhengzhou (\begin{CJK*}{UTF8}{gbsn}到郑州最晚的车次是几点\end{CJK*}). We also provide a English version of twenty three cases in every domain in Fig \ref{fig:english domain case}.

\section{Broader Impact and Ethnic Consideration}

Data in DialogUSR does not involve user privacy. The data source we collect from SMP-ECDT and RiSAWOZ is open source for research and is licensed under the MIT License which is a short and simple permissive license with conditions only requiring preservation of copyright and license notices. 

Our generative baseline models have very low risk in terms of producing discriminatory, insulting words or divulging privacy due to the fact all the training data are strictly screened and do not include private user information or insulting content. All involved annotators voluntarily participated with decent payment.

\clearpage
\newpage

\end{document}